\documentclass[sigconf]{acmart}

\usepackage{booktabs}
\usepackage{caption}
\usepackage{subfig}
\usepackage{times}
\usepackage{graphicx}
\usepackage{amssymb,amsmath,amsfonts}

\setcopyright{none}
\acmConference[SysML'18]{SysML Conference}{February 2018}{Stanford, CA, USA}
\acmYear{2018}
\acmDOI{}
\acmISBN{}

\begin{document}
\title{Greenhouse: A Zero-Positive Machine Learning System \\ for Time-Series Anomaly Detection}

\author{Tae Jun Lee}
\authornote{The work was done while a Brown student.}
\affiliation{
  \institution{Microsoft}
  %\city{Redmond}
  %\state{WA}
}
%\email{tae_jun_lee@alumni.brown.edu}

\author{Justin Gottschlich, Nesime Tatbul}
\affiliation{
  \institution{Intel Labs}
  %\city{Santa Clara}
  %\state{CA}
}
%\email{{justin.gottschlich, nesime.tatbul}@intel.com}

\author{Eric Metcalf, Stan Zdonik}
\affiliation{
  \institution{Brown University}
  %\city{Providence}
  %\state{RI}
}
%\email{{emetcalf, sbz}@cs.brown.edu}

\renewcommand{\shortauthors}{T. J. Lee, J. Gottschlich, N. Tatbul, E. Metcalf, S. Zdonik}

\begin{abstract}
This short paper describes our ongoing research on Greenhouse - a zero-positive machine learning system for time-series anomaly detection.
\end{abstract}

\maketitle

\section{Introduction} \label{sec:intro}

The emerging ``killer app'' of the Internet of Things (IoT) envisions
a world made up of huge numbers of sensors and computing devices that
interact in intelligent ways (e.g., self-driving cars, industrial automation, mobile phone tracking). The sensors produce massive amounts of data and the computing devices must figure out how to use it. The preponderance of this data is numerical time-series.

Anomaly detection, i.e., the process of finding patterns that do not conform to expected behavior, over time-series is an important capability in IoT with multiple potential applications. Through anomaly detection, we can identify unusual environmental situations that need human attention \cite{macrobase}, distinguish outliers in sensor data cleaning \cite{outlier}, or pre-filter uninteresting portions of data to preserve computing resources, to name a few.

\begin{comment}
\begin{itemize}

\item {\bf Identifying unusual events:} 
Many IoT applications monitor their environment to detect situations that 
need attention from either a human or another computing device. Such 
applications are looking for anomalous patterns in the time-series data.

\item {\bf Filtering uninteresting data:}
Large-scale applications such as autonomous vehicles produce GBs of data per vehicle per day \cite{oxford-dataset}. It is a challenge to store and transfer all of this data, yet typically only a small fraction of it is really interesting for long term use. Anomaly detection can serve as a filter to determine which portion of the data is worth keeping.

\item {\bf Cleaning noisy data:}
IoT data can be lossy or erroneous due to environmental factors, misbehaving devices, etc. Being able to detect outlier readings is a key first step in dealing with such data.

\end{itemize}
\end{comment}

Prior research on time-series anomaly detection largely relied on traditional data mining and machine learning techniques \cite{anomaly-survey}. More recently, new techniques using deep neural networks have gained attention. Recurrent neural networks (RNNs), and in particular their Long Short-Term Memory (LSTM) variant, excel at capturing short-term dependencies when  making predictions over sequence data \cite{lstm, lstm-blog}. Specifically, LSTM has the ability to support new data as well as a way to gracefully forget old, and therefore less relevant, data. As such, it is a good fit for predicting time-varying patterns over sequential data, as in time-series anomaly detection \cite{malhotra:2015:esann}.

In this paper, we describe a novel time-series anomaly detection system called Greenhouse. Our key goal in Greenhouse is to combine state-of-the-art machine learning and data management techniques for efficient and accurate prediction of anomalous patterns over high volumes of time-series data. We have designed Greenhouse as a \emph{zero-positive} machine learning system \cite{autoperf}, in that, it does not require anomalous samples in its training datasets. This capability has notable practical value, because it can be challenging to collect and label anomalous data when anomalies are rare and varying.

In what follows, we first summarize our algorithmic machine learning framework and preliminary results from its implementation on top of the TensorFlow machine learning library \cite{tensorflow-osdi16}. Then we discuss our ongoing research agenda in extending this framework into an end-to-end time-series anomaly detection system for IoT.

\begin{comment}
In our preliminary work to date, we have built an LSTM-based time-series
anomaly detection system called Greenhouse. Greenhouse is built on top of the Apache TensorFlow machine learning library
\cite{tensorflow} using the Keras neural network library extension \cite{keras}.
The key aspects of our system design are:

\begin{itemize}

\item Greenhouse is a zero-positive / zero-shot learning approach in that
it does not require anomalous samples to exist in the training datasets.
This capability has significant practical value, since it is challenging to collect and label anomalous data when anomalies can be rare and varying.

\item Greenhouse can detect contextual and collective anomalies (a.k.a., range-based anomalies)  \cite{anomaly-survey}, which makes it more powerful than point-based anomaly detection approaches.

\end{itemize}
\end{comment}

\section{Algorithmic Framework} \label{sec:algorithmic}

At the heart of Greenhouse's algorithmic framework is an LSTM-based neural network model, which is used to predict values at future time points based on values observed at past time points. This model is built based on a training dataset which represents what is considered to be ``normal'', and is then used for detecting anomalous values - those that sufficiently deviate from what the model predicts would be normally observed.

%We have designed Greenhouse as a ``zero-positive" machine learning system in that, it does not require anomalous samples to exist in its training datasets. This capability has significant practical value, since it is challenging to collect and label anomalous data when anomalies can be rare and varying.
%Greenhouse can detect contextual and collective anomalies (a.k.a., range-based anomalies)  \cite{anomaly-survey}, which makes it more powerful than point-based anomaly detection approaches.

Given a time-series that consists of an isochronous (i.e., evenly spaced in time) sequence of (time, value) pairs as illustrated in Table \ref{tab:GH-example}, there are two basic tasks that are repeatedly performed in our framework:

\noindent
{\bf Making a prediction:}
For a given time point $t$, a window of most recently observed values [$v_{t-B}, .., v_{t-1}$] of length $B$ is used as ``Look-Back'' to predict a subsequent window of values [$v_{t}, .., v_{t+F-1}$] of length $F$ as ``Predict-Forward''. This is applied to all points in the given time-series in a sliding window fashion, resulting in $F$ distinct value predictions for each time point. For example, in Table \ref{tab:GH-prediction}, for $t=4$, a Look-Back window with observed values [$v_1, v_2, v_3$] is used to predict a Predict-Forward window with values [$p_{4.1}, p_{5.1}$]. As a result, we obtain two value predictions per time point (except the initial point), e.g., $p_{5.1}$ and $p_{5.2}$ for $t=5$.

\noindent
{\bf Computing an error vector:}
For a given time point $t$, an error vector ``Error Vector($t$)'' of length $F$ is computed. This vector consists of differences between predicted and observed values that correspond to time point $t$. For example, in Table \ref{tab:GH-error}, the error vector for $t=5$ consists of two values: [$p_{5.1} - v_5, p_{5.2} - v_5$]. 
%As an example, Table \ref{tab:GH-example} illustrates how to make predictions and compute error vectors for a given time-series of (time, value) pairs: $[ (1, v_1), (2, v_2), (3, v_3), (4, v_4), (5, v_5), ... ]$.

\begin{table}[h]
\small
\begin{center}
\caption{Predictions and Error Vectors for an example time-series (time, value) = [ $(1, v_1)$, $(2, v_2)$, $(3, v_3)$, $(4, v_4)$, $(5, v_5)$, ... ]}
\label{tab:GH-example}
\subfloat[Making a Prediction]{
\begin{tabular}{cc}
\toprule
Look-Back & Predict-Forward \\
    (B=3) & (F=2) \\
\midrule
$v_1, v_2, v_3$ & $p_{4.1}, p_{5.1}$ \\
$v_2, v_3, v_4$ & $p_{5.2}, p_{6.1}$ \\
$v_3, v_4, v_5$ & $p_{6.2}, p_{7.1}$ \\
$v_4, v_5, v_6$ & $p_{7.2}, p_{8.1}$ \\
... & ... \\
\bottomrule
\end{tabular}
\label{tab:GH-prediction}
}
\hspace{0.3cm}
%\\
\subfloat[Computing an Error Vector]{
\begin{tabular}{cc}
\toprule
Time & Error \\
Point & Vector (t) \\
\midrule
t = 5 & $[p_{5.1} - v_5, p_{5.2} - v_5]$ \\
t = 6 & $[p_{6.1} - v_6, p_{6.2} - v_6]$ \\
t = 7 & $[p_{7.1} - v_7, p_{7.2} - v_7]$ \\
t = 8 & $[p_{8.1} - v_8, p_{8.2} - v_8]$ \\
... & ... \\
\bottomrule
\end{tabular}
\label{tab:GH-error}
}
\end{center}
\end{table}

\begin{figure}[t]
\centering
\includegraphics[width=\columnwidth]{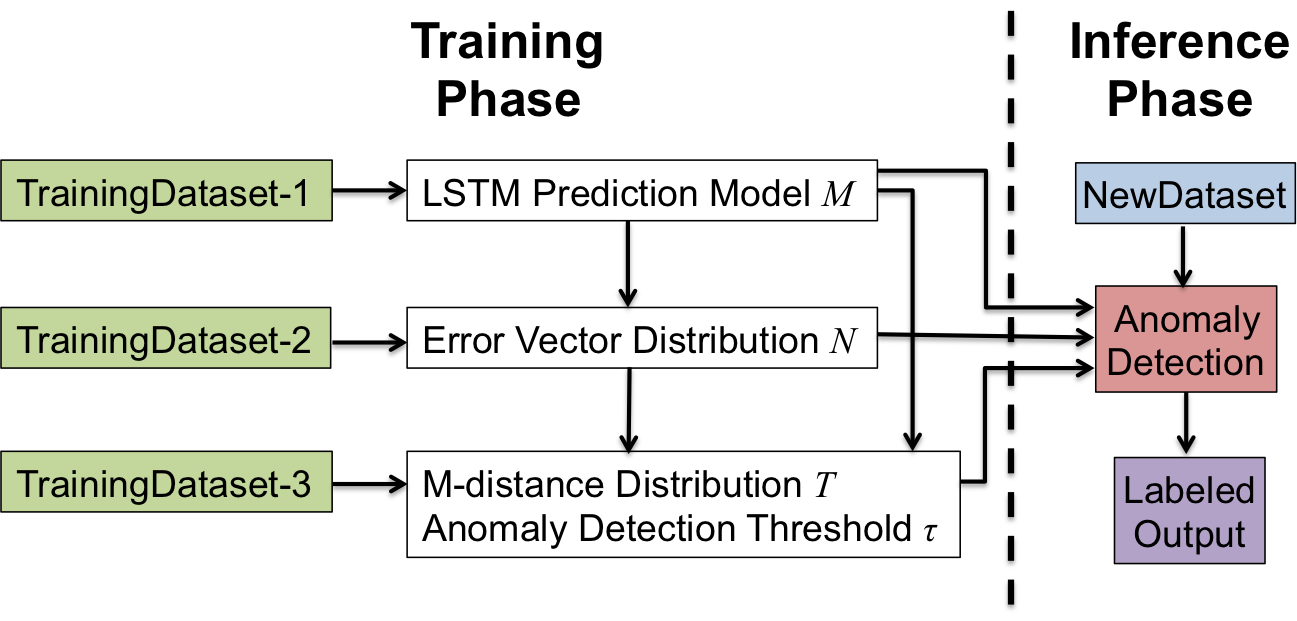}
\caption{Training and Inference Phases in Greenhouse}
\label{fig:GH-workflow}
\end{figure}

\vspace{-0.2in}
As illustrated in Figure \ref{fig:GH-workflow}, our overall framework mainly consists of two phases:
(i) build an LSTM model ($M$) and a multi-variate probability distribution for error vectors ($N$), and determine an error threshold for anomalies ($\tau$) during a {\em Training Phase};
(ii) use $M$, $N$, and $\tau$ to detect anomalies over previously unseen data sets during an {\em Inference Phase}. We now provide a step-by-step summary of these two phases.

\noindent
{\bf Training Phase:} Training is performed in four key steps:

\begin{enumerate}

\item Split a non-anomalous dataset into three: TrainingDataset-1, TrainingDataset-2, TrainingDataset-3.

\item Train an LSTM prediction model $M$ with TrainingDataset-1.

\item Apply $M$ over TrainingDataset-2 to make predictions and compute error vectors. Then fit the resulting error vectors into a multi-variate normal distribution $N$.

\item Apply $M$ over TrainingDataset-3 to make predictions and compute error vectors. Then compute M-distances (Mahalano\-bis-distances \cite{mahalanobis}), and fit the resulting M-distances into a truncated normal distribution $T$. Finally, evaluate the inverse cumulative distribution function of $T$ at a user-specified percentile to be used as the anomaly detection threshold $\tau$.

\end{enumerate}

\noindent
{\bf Inference Phase:} Inference involves the following steps:

\begin{enumerate}

\item Apply $M$ over NewDataset to make predictions.

\item Compute error vectors.

\item Compute M-distances between these error vectors and the center of $N$.

\item Finally, label the time-series values whose M-distances exceed $\tau$ as anomalies.

\end{enumerate}

\begin{table}[h]
\centering
\caption{Greenhouse vs. LSTM-AD \cite{malhotra:2015:esann}}
\label{tab:GH-result}
\begin{tabular}{lccc}
\toprule
& Precision & Recall & $F_1$ score \\
\midrule
Greenhouse (Twitter\_AAPL) & 0.49 & 0.06 & 0.11 \\
LSTM-AD (Twitter\_AAPL) & 0.22 & 0.14 & 0.17 \\
Greenhouse (nyc\_taxi) & 0.25 & 0.58 & 0.35 \\
LSTM-AD (nyc\_taxi) & 0.26 & 0.82 & 0.40 \\
\bottomrule
\end{tabular}
\end{table}

We implemented this algorithm on top of TensorFlow \cite{tensorflow-osdi16}, and compared its anomaly detection accuracy with LSTM-AD \cite{malhotra:2015:esann}, based on real-world datasets from the Numenta Anomaly Benchmark \cite{lavin:anomaly:2015}. LSTM-AD fundamentally differs from Greenhouse in that, it is not zero-positive, i.e., it requires anomalous samples to be present in its training datasets. We present a sample experimental result in Table \ref{tab:GH-result}.
For two different datasets, Greenhouse  performed favorably over LSTM-AD in terms of its Precision and managed to  maintain a close $F_1$ score, despite its lower Recall. This is a remarkable result, especially given that Greenhouse uses significantly smaller training data (about 25\% and 55\% that of LSTM-AD, respectively) and does not rely on any anomalous samples. We believe these are powerful qualities, making Greenhouse more generally applicable in practice.
%Unlike our zero-positive approach, LSTM-AD requires {\em larger} training datasets which contain {\em anomalous samples}. %Therefore, it is less effective at detecting anomalies that are not similar to the ones in the training samples, while Greenhouse does not have this restriction. Furthermore, the size of LSTM-AD's training dataset tends to be much greater than that of Greenhouse's, since it requires both anomalous and non-anomalous data.
%As can be seen, despite using sigfniciantly smaller training data (\~25\% of LSTM-AD's train dataset in exp1 and \~55\% in exp2)  Greenhouse managed to perform close to LSTM-AD in terms of F1-score, and outperformed it in terms of precision score in exp1.
%when tested on two separate real-world datasets (exp1 and exp2) from the Numenta Anomaly Benchmark \cite{lavin:anomaly:2015}.

\section{Ongoing Research} \label{sec:research}

Going forward, we are working on a number of research issues to extend our algorithmic framework into a full-feature time-series anomaly detection system for IoT. We conclude the paper with an overview of these issues.

\noindent
{\bf Training dataset management:} In Greenhouse, we use multiple datasets during training. Furthermore, for sequential data like time-series, respecting order, regularity (isochronism), and continuity is important in correctly capturing patterns of interest. In order to avoid the risk of over/under-fitting and to preserve continuity, we need to pay attention to how we choose and arrange our training datasets. 

\noindent
{\bf Range-based anomalies and their evaluation:} Time-series anomalies often manifest themselves over a period of time rather than at single time points (so-called range-based or collective anomalies \cite{anomaly-survey, collective}). Furthermore, judging accuracy of results in this context is highly intricate and application-dependent. We are extending Greenhouse with models and algorithms to handle range-based anomalies in a principled manner \cite{accuracy-sysml18}.

\noindent
{\bf Real-time anomaly detection:} Our algorithmic framework initially focused on operating in an ``offline mode'', where both training and inference are applied on finite datasets that have been collected in the past and are being analyzed retrospectively after the fact. In many IoT applications, real-time anomaly detection on live data is also important. Thus, we are extending Greenhouse to operate in an ``online mode'', which requires continuous, low-latency inferencing (and possibly training) over streaming time-series. This includes things like carefully analyzing the impact of algorithm and model parameters on accuracy and performance, and properly tuning them as datasets change.

\noindent
{\bf Utilizing human feedback:} When deployed in an online setting, Greenhouse will start making anomaly predictions on new datasets as well as potentially saving these (self-labeled) datasets for further training. Furthermore, feedback from human may be available to validate anomaly predictions or to course-correct. We plan to augment Greenhouse with reinforcement learning techniques to utilize such feedback when available \cite{rl-lstm}.

\noindent
{\bf Data management support:} As in any deep learning system, data is an indispensable resource in Greenhouse. We will deploy Greenhouse and study its data management needs within the context of a time-series data management system environment (e.g., Metronome \cite{metronome}). This includes things like providing methods of efficient and consistent access to arbitrary windows of data needed by Greenhouse algorithms.

\noindent
{\bf Exploiting high-performance compute frameworks:} While implemented on top of a state-of-the-art machine learning framework \cite{tensorflow-osdi16}, Greenhouse does not yet take full advantage of high-performance compute facilities provided by such frameworks. We plan to extend Greenhouse in this direction, including exploring its use in conjunction with other complementary tools and libraries, such as Intel\textsuperscript{\textregistered} Nervana\textsuperscript{\texttrademark} Graph \cite{nervana}.

\noindent
{\bf Support for distributed IoT deployments:} IoT applications typically execute on multi-tier deployments from wireless networks of edge devices to more powerful cloud servers. We plan to investigate the use of Greenhouse in such distributed deployments with heterogeneous computing resources.

\vspace{0.05in}
\noindent
{\bf Acknowledgments.}
This research has been funded in part by Intel.

\newpage

\bibliographystyle{ACM-Reference-Format}
\bibliography{greenhouse}


\begin{thebibliography}{15}

%%% the Web address.

\ifx \showCODEN    \undefined \def \showCODEN     #1{\unskip}     \fi
\ifx \showDOI      \undefined \def \showDOI       #1{#1}\fi
\ifx \showISBNx    \undefined \def \showISBNx     #1{\unskip}     \fi
\ifx \showISBNxiii \undefined \def \showISBNxiii  #1{\unskip}     \fi
\ifx \showISSN     \undefined \def \showISSN      #1{\unskip}     \fi
\ifx \showLCCN     \undefined \def \showLCCN      #1{\unskip}     \fi
\ifx \shownote     \undefined \def \shownote      #1{#1}          \fi
\ifx \showarticletitle \undefined \def \showarticletitle #1{#1}   \fi
\ifx \showURL      \undefined \def \showURL       {\relax}        \fi
% invisible to TeX
\providecommand\bibfield[2]{#2}
\providecommand\bibinfo[2]{#2}
\providecommand\natexlab[1]{#1}
\providecommand\showeprint[2][]{arXiv:#2}

\bibitem[\protect\citeauthoryear{Abadi, Barham, Chen, Chen, Davis, Dean, Devin,
  Ghemawat, Irving, Isard, Kudlur, Levenberg, Monga, Moore, Murray, Steiner,
  Tucker, Vasudevan, Warden, Wicke, Yu, and Zheng}{Abadi et~al\mbox{.}}{2016}]%
        {tensorflow-osdi16}
\bibfield{author}{\bibinfo{person}{Martin Abadi}, \bibinfo{person}{Paul
  Barham}, \bibinfo{person}{Jianmin Chen}, \bibinfo{person}{Zhifeng Chen},
  \bibinfo{person}{Andy Davis}, \bibinfo{person}{Jeffrey Dean},
  \bibinfo{person}{Matthieu Devin}, \bibinfo{person}{Sanjay Ghemawat},
  \bibinfo{person}{Geoffrey Irving}, \bibinfo{person}{Michael Isard},
  \bibinfo{person}{Manjunath Kudlur}, \bibinfo{person}{Josh Levenberg},
  \bibinfo{person}{Rajat Monga}, \bibinfo{person}{Sherry Moore},
  \bibinfo{person}{Derek~G. Murray}, \bibinfo{person}{Benoit Steiner},
  \bibinfo{person}{Paul Tucker}, \bibinfo{person}{Vijay Vasudevan},
  \bibinfo{person}{Pete Warden}, \bibinfo{person}{Martin Wicke},
  \bibinfo{person}{Yuan Yu}, {and} \bibinfo{person}{Xiaoqiang Zheng}.}
  \bibinfo{year}{2016}\natexlab{}.
\newblock \showarticletitle{{TensorFlow: A System for Large-scale Machine
  Learning}}. In \bibinfo{booktitle}{\emph{{12th USENIX Symposium on Operating
  Systems Design and Implementation (OSDI)}}}. \bibinfo{pages}{265--283}.
\newblock


\bibitem[\protect\citeauthoryear{Aggarwal}{Aggarwal}{2013}]%
        {outlier}
\bibfield{author}{\bibinfo{person}{Charu~C. Aggarwal}.}
  \bibinfo{year}{2013}\natexlab{}.
\newblock \bibinfo{booktitle}{\emph{{Outlier Analysis}}}.
\newblock \bibinfo{publisher}{Springer}.
\newblock


\bibitem[\protect\citeauthoryear{Alam, Gottschlich, and Muzahid}{Alam
  et~al\mbox{.}}{2017}]%
        {autoperf}
\bibfield{author}{\bibinfo{person}{Mohammad Mejbah~Ul Alam},
  \bibinfo{person}{Justin Gottschlich}, {and} \bibinfo{person}{Abdullah
  Muzahid}.} \bibinfo{year}{2017}\natexlab{}.
\newblock \bibinfo{booktitle}{\emph{{AutoPerf: A Generalized Zero-Positive
  Learning System to Detect Software Performance Anomalies}}}.
\newblock \bibinfo{type}{{T}echnical {R}eport}.
\newblock
\urldef\tempurl%
\url{http://arxiv.org/abs/1709.07536/}
\showURL{%
\tempurl}


\bibitem[\protect\citeauthoryear{Bailis, Gan, Madden, Narayanan, Rong, and
  Suri}{Bailis et~al\mbox{.}}{2017}]%
        {macrobase}
\bibfield{author}{\bibinfo{person}{Peter Bailis}, \bibinfo{person}{Edward Gan},
  \bibinfo{person}{Samuel Madden}, \bibinfo{person}{Deepak Narayanan},
  \bibinfo{person}{Kexin Rong}, {and} \bibinfo{person}{Sahaana Suri}.}
  \bibinfo{year}{2017}\natexlab{}.
\newblock \showarticletitle{{MacroBase: Prioritizing Attention in Fast Data}}.
  In \bibinfo{booktitle}{\emph{{ACM International Conference on Management of
  Data (SIGMOD)}}}. \bibinfo{pages}{541--556}.
\newblock


\bibitem[\protect\citeauthoryear{Bakker}{Bakker}{2001}]%
        {rl-lstm}
\bibfield{author}{\bibinfo{person}{Bram Bakker}.}
  \bibinfo{year}{2001}\natexlab{}.
\newblock \showarticletitle{{Reinforcement Learning with Long Short-term
  Memory}}. In \bibinfo{booktitle}{\emph{{14th International Conference on
  Neural Information Processing Systems (NIPS)}}}. \bibinfo{pages}{1475--1482}.
\newblock


\bibitem[\protect\citeauthoryear{Bontemps, Cao, McDermott, and
  Le-Khac}{Bontemps et~al\mbox{.}}{2017}]%
        {collective}
\bibfield{author}{\bibinfo{person}{Loic Bontemps}, \bibinfo{person}{Van~Loi
  Cao}, \bibinfo{person}{James McDermott}, {and} \bibinfo{person}{Nhien-An
  Le-Khac}.} \bibinfo{year}{2017}\natexlab{}.
\newblock \bibinfo{booktitle}{\emph{{Collective Anomaly Detection based on Long
  Short Term Memory Recurrent Neural Network}}}.
\newblock \bibinfo{type}{{T}echnical {R}eport}.
\newblock
\urldef\tempurl%
\url{https://arxiv.org/abs/1703.09752}
\showURL{%
\tempurl}


\bibitem[\protect\citeauthoryear{Chandola, Banerjee, and Kumar}{Chandola
  et~al\mbox{.}}{2009}]%
        {anomaly-survey}
\bibfield{author}{\bibinfo{person}{Varun Chandola}, \bibinfo{person}{Arindam
  Banerjee}, {and} \bibinfo{person}{Vipin Kumar}.}
  \bibinfo{year}{2009}\natexlab{}.
\newblock \showarticletitle{{Anomaly Detection: A Survey}}.
\newblock \bibinfo{journal}{\emph{{ACM Computing Surveys}}}
  \bibinfo{volume}{41}, \bibinfo{number}{3} (\bibinfo{year}{2009}),
  \bibinfo{pages}{15:1--15:58}.
\newblock


\bibitem[\protect\citeauthoryear{{Christopher Olah}}{{Christopher
  Olah}}{2015}]%
        {lstm-blog}
\bibfield{author}{\bibinfo{person}{{Christopher Olah}}.}
  \bibinfo{year}{2015}\natexlab{}.
\newblock \bibinfo{title}{{Understanding LSTM Networks}}.
\newblock \bibinfo{howpublished}{\footnotesize
  \url{http://colah.github.io/posts/2015-08-Understanding-LSTMs/}}.
  (\bibinfo{year}{2015}).
\newblock


\bibitem[\protect\citeauthoryear{Hochreiter and Schmidhuber}{Hochreiter and
  Schmidhuber}{1997}]%
        {lstm}
\bibfield{author}{\bibinfo{person}{Sepp Hochreiter} {and}
  \bibinfo{person}{Jurgen Schmidhuber}.} \bibinfo{year}{1997}\natexlab{}.
\newblock \showarticletitle{{Long Short-Term Memory}}.
\newblock \bibinfo{journal}{\emph{{Neural Computation}}} \bibinfo{volume}{9},
  \bibinfo{number}{8} (\bibinfo{year}{1997}), \bibinfo{pages}{1735--1780}.
\newblock


\bibitem[\protect\citeauthoryear{Intel}{Intel}{2017}]%
        {nervana}
\bibfield{author}{\bibinfo{person}{Intel}.} \bibinfo{year}{2017}\natexlab{}.
\newblock \bibinfo{title}{{Intel Nervana Graph}}.
\newblock \bibinfo{howpublished}{\footnotesize
  \url{http://ngraph.nervanasys.com/}}.   (\bibinfo{year}{2017}).
\newblock


\bibitem[\protect\citeauthoryear{Lavin and Ahmad}{Lavin and Ahmad}{2015}]%
        {lavin:anomaly:2015}
\bibfield{author}{\bibinfo{person}{Alexander Lavin} {and}
  \bibinfo{person}{Subutai Ahmad}.} \bibinfo{year}{2015}\natexlab{}.
\newblock \showarticletitle{{Evaluating Real-Time Anomaly Detection Algorithms
  - The Numenta Anomaly Benchmark}}. In \bibinfo{booktitle}{\emph{{14th IEEE
  International Conference on Machine Learning and Applications (ICMLA)}}}.
  \bibinfo{pages}{38--44}.
\newblock


\bibitem[\protect\citeauthoryear{Lee, Gottschlich, Tatbul, Metcalf, and
  Zdonik}{Lee et~al\mbox{.}}{2018}]%
        {accuracy-sysml18}
\bibfield{author}{\bibinfo{person}{Tae~Jun Lee}, \bibinfo{person}{Justin
  Gottschlich}, \bibinfo{person}{Nesime Tatbul}, \bibinfo{person}{Eric
  Metcalf}, {and} \bibinfo{person}{Stan Zdonik}.}
  \bibinfo{year}{2018}\natexlab{}.
\newblock \showarticletitle{{Precision and Recall for Range-Based Anomaly
  Detection}}. \bibinfo{howpublished}{\footnotesize
  \url{https://arxiv.org/abs/1801.03175/}}. In \bibinfo{booktitle}{\emph{{SysML
  Conference}}}.
\newblock


\bibitem[\protect\citeauthoryear{Mahalanobis}{Mahalanobis}{1936}]%
        {mahalanobis}
\bibfield{author}{\bibinfo{person}{Prasanta~Chandra Mahalanobis}.}
  \bibinfo{year}{1936}\natexlab{}.
\newblock \showarticletitle{{On the Generalised Distance in Statistics}}.
\newblock \bibinfo{journal}{\emph{{Proceedings of the National Institute of
  Sciences of India}}} \bibinfo{volume}{2}, \bibinfo{number}{1}
  (\bibinfo{year}{1936}), \bibinfo{pages}{49--55}.
\newblock


\bibitem[\protect\citeauthoryear{Malhotra, Vig, Shroff, and Agarwal}{Malhotra
  et~al\mbox{.}}{2015}]%
        {malhotra:2015:esann}
\bibfield{author}{\bibinfo{person}{Pankaj Malhotra}, \bibinfo{person}{Lovekesh
  Vig}, \bibinfo{person}{Gautam Shroff}, {and} \bibinfo{person}{Puneet
  Agarwal}.} \bibinfo{year}{2015}\natexlab{}.
\newblock \showarticletitle{{Long Short Term Memory Networks for Anomaly
  Detection in Time Series}}. In \bibinfo{booktitle}{\emph{{European Symposium
  on Artificial Neural Networks, Computational Intelligence and Machine
  Learning (ESANN)}}}. \bibinfo{pages}{89--94}.
\newblock


\bibitem[\protect\citeauthoryear{Meehan, Aslantas, Zdonik, Tatbul, and
  Du}{Meehan et~al\mbox{.}}{2017}]%
        {metronome}
\bibfield{author}{\bibinfo{person}{John Meehan}, \bibinfo{person}{Cansu
  Aslantas}, \bibinfo{person}{Stan Zdonik}, \bibinfo{person}{Nesime Tatbul},
  {and} \bibinfo{person}{Jiang Du}.} \bibinfo{year}{2017}\natexlab{}.
\newblock \showarticletitle{{Data Ingestion for the Connected World}}. In
  \bibinfo{booktitle}{\emph{{Conference on Innovative Data Systems Research
  (CIDR)}}}.
\newblock


\end{thebibliography}

\end{document}